# Transforming Prioritized Defaults and Specificity into Parallel Defaults

**Benjamin N. Grosof**
IBM   T. J. Watson Research Center
P.O. Box 704, Yorktown Heights, NY 10598
(914) 784-7783 direct -7455 fax -7100 main
Internet: grosof@watson.ibm.com (alt.: grosof@cs.stanford.edu)

## Abstract

We show how to transform any set of prioritized propositional defaults into an equivalent set of parallel (i.e., unprioritized) defaults, in circumscription. We give an algorithm to implement the transform. We show how to use the transform algorithm as a generator of a whole family of inferencing algorithms for circumscription. The method is to employ the transform algorithm as a front end to any inferencing algorithm, e.g., one of the previously available, that handles the parallel (empty) case of prioritization. Our algorithms provide not just coverage of a new expressive class, but also alternatives to previous algorithms for implementing the previously covered class (layered) of prioritization.

In particular, we give a new query-answering algorithm for prioritized cirumscription which is sound and complete for the full expressive class of unrestricted finite prioritization partial orders, for propositional defaults (or minimized predicates). By contrast, previous algorithms required that the prioritization partial order be *layered*, i.e., structured similar to the system of rank in the military.

Our algorithm enables, for the first time, the implementation of the most useful class of prioritization: *non*-layered prioritization partial orders. Default inheritance, for example, typically requires non-layered prioritization to represent specificity adequately. Our algorithm enables not only the implementation of default inheritance (and specificity) within prioritized circumscription, but also the extension and combination of default inheritance with other kinds of prioritized default reasoning, e.g.: with stratified logic programs with negation-as-failure. Such logic programs are previously known to be representable equivalently as layered-priority predicate circumscriptions.

Worst-case, the transform increases the number of defaults exponentially. We discuss how inferencing is practically implementable nevertheless in two kinds of situations: general expressiveness but small numbers of defaults, or expressive special cases with larger numbers of defaults. One such expressive special case is *non-top-heaviness* of the prioritization partial order.

In addition to its direct implementation, the transform can also be exploited analytically to generate special case algorithms, e.g., a tractable transform for a class within default inheritance (detailed in another, forthcoming paper).

We discuss other aspects of the significance of the fundamental result. One can view the transform as reducing $n$ degrees of partially ordered belief confidence to just 2 degrees of confidence: for-sure and (unprioritized) default. Ordinary, parallel default reasoning, e.g., in parallel circumscription or Poole's Theorist, can be viewed in these terms as reducing 2 degrees of confidence to just 1 degree of confidence: that of the non-monotonic theory's conclusions. The expressive reduction's computational complexity suggests that prioritization is valuable for its expressive conciseness, just as defaults are for theirs.

For Reiter's Default Logic and Poole's Theorist, the transform implies how to extend those formalisms so as to equip them with a concept of prioritization that is exactly equivalent to that in circumscription. This provides an interesting alternative to Brewka's approach to equipping them with prioritization-type precedence.

*A longer version of this paper, including the proof of the central transform result, will soon be available as an IBM Research Report.*

## 1 INTRODUCTION

**THE IMPORTANCE OF PRIORITIZATION-TYPE PRECEDENCE:**
Prioritization-type precedence is an important aspect of default and non-monotonic reasoning. It is a widely studied concept (actually, family of related concepts) of multiple degrees of confidence (uncertainty, reliability) in beliefs: partially ordered, and lexicographic in flavor. By prioritization-type "precedence", we mean an override relationship between defaults that governs the resolution of conflicts between defaults; so that, in case of pairwise conflicts between defaults, a higher-precedence default "wins", i.e., goes through as non-monotonically entailed, in case of conflict with a lower-precedence default. Each default can be viewed as an uncertain belief.

Prioritization in circumscription [McCarthy, 1986] [Lifschitz, 1985] [Grosof, 1991] is one of the first, and most expressively general, concepts of precedence in the default reasoning literature.

Several other non-monotonic formalisms share a somewhat similar concept of prioritization-type precedence / degrees of confidence (and a somewhat similar concept of a default) to that in circumscription, including: the specificity dominance principle employed in inheritance, e.g., cf. [Touretzky, 1986] [Stein, 1989] [Quantz and Royer, 1992] [Geffner, 1992], conditional logics, e.g., cf. [Delgrande, 1987a] [Delgrande, 1987b] [Geffner, 1992], and argument systems, e.g., cf. [Loui, 1987]; aggregation principles for model-preference logics [Brown and Shoham, 1989], including for logic programming with negation-as-failure [Przymusinski, 1988] and terminological logics [Quantz and Royer, 1992]; possibilistic logic [Dubois and Prade, 1988]; syntax-based belief revision formalisms, e.g., [Nebel, 1989]; and a variety of others, e.g., [Brewka, 1989a] [Brewka, 1989b] [Brewka, 1994] [Ginsberg, 1988] [Zadrozny, 1987] [Pollock, 1987] [Konolige, 1988] [Ryan, 1992b] [Ryan, 1992a] [Hunter, 1994].

From a practical viewpoint, two very important kinds of non-monotonic reasoning, in use even before the knowledge representation (KR) field of non-monotonic logic started in the late 1970's, are:
1. inheritance with exceptions, e.g., default inheritance in frame-based KR systems; and
2. use of negation-as-failure in logic programming, e.g., in Prolog.

Both these kinds of non-monotonic reasoning can be viewed formally in terms of defaults. In both, prioritization-type precedence is then very important to represent desired behavior: in default inheritance, to represent specificity dominance (i.e., the precedence of a default with more specific-class antecedent over a default with a more general-class antecedent); and, in stratified logic programs, to represent recursive depth in negation-as-failure use (i.e., deeper strata in backward inferencing have higher precedence associated with their predicates' minimization) [Lifschitz, 1987] [Przymusinski, 1988].

More generally, precedence appears to be an important expressive aspect of defaults needed or useful to represent many domains. Bases for precedence information include not only specificity dominance but also reliability and authority of sources [Grosof, 1993], decision-theoretic utility (e.g., rules about emergencies have higher precedence) [Grosof, 1991] [Poole, 1992], and temporal directionality (e.g., freshness; or Shoham's [1988] chronological minimization).

## VIRTUES OF CIRCUMSCRIPTION AS A FOCUS FORMALISM:

Circumscription is a historically central and relatively well-studied non-monotonic formalism, with a number of attractive characteristics. First, its expressive feature of prioritization can directly represent precedence; and do so, moreover, in a relatively expressively powerful fashion. By contrast, many other non-monotonic formalisms do not have any expressive feature to directly represent precedence: e.g., Default Logic [Reiter, 1980] and Autoepistemic Logic [Moore, 1985]. Second, circumscription captures a core notion of default shared by most formalisms, corresponding closely to that of Poole's Theorist formalism [1988]. Third, circumscription is skeptical, which we find to be typically more useful for applications than brave. Fourth, circumscription has a relatively attractive model theory. Fifth, much is known about relationships between circumscription and a variety of other non-monotonic formalisms, including Default Logic and Autoepistemic Logic.

In particular, prioritized (default) circumscription has been previously shown to be able to represent the stratified class of logic programs with negation-as-failure. This is a large, interesting class that is important in practical implementations and applications of logic programming. Prioritized circumscription has also been previously shown to be able to represent a large, interesting class of default inheritance.

We observe that prioritized (default) cirumscription can, moreover, represent theories that combine default inheritance and logic programs, as well as that extend them expressively.

Previously, however, there have not been any query-answering inferencing algorithms for sufficiently general cases of prioritization: i.e., to handle prioritization partial orders that are not *layered* (a.k.a. stratified), in the sense discussed first in [Grosof, 1991] and studied more in [Grosof, 1992b]. In particular, as discussed there, non-layered prioritization partial orders are required to adequately represent even very simple cases of specificity in default inheritance. Layered prioritization cannot adequately represent default inheritance, much less its combination and extension with logic programs.

(Sections 2 and 3 discuss details of how circumscription represents these classes.)

## PROBLEMS AND OPPORTUNITIES ADDRESSED:

Our overall, primary motivations in this paper are twofold. The first is to advance fundamental understanding of prioritization-type precedence as a kind of partially-ordered degree of confidence (uncertainty, reliability) among beliefs. The second is to advance the implementation of default reasoning that has prioritization-type precedence.

More particularly, we are interested in prioritized cir-cumscription, in great part because of its ability to represent, to a considerable extent, the combination and extension of logic programs with negation-as-failure and default inheritance.

The prospect of practically implementing such combination and extension is an exciting opportunity. It could provide closer integration and enhancement of two very important kinds of non-monotonic reasoning that are today in practical use as basic programming mechanisms.

In this paper, we accomplish (among other things) a first step towards this vision: to provide a correct algorithm for query-answering inference, where previously there was no algorithm at all. More precisely, we give a family of sound and complete query-answering algorithms for the full class of circumscriptions in which the prioritization partial order is unrestricted, and the defaults are propositional. This expressive class suffices to include the combination and extension of large classes of logic programs and default inheritance.

## 2   PRELIMINARY DEFINITIONS

Below, and throughout the paper, we follow the definitions, notation, and terminology of [Grosof, 1991] and of [Grosof, 1992b]. Refer to either the former (sections 3 and 4 there) or the latter (sections 2.2, 2.3, 2.7, 2.9 there) to find what is not elaborated below.

A *pre-order* is a transitive, reflexive binary relation; i.e., a kind of ordering. A *default* pre-order $\preceq_{Di}$ is one that expresses the preference to maximize a first-order formula $Di$ (in predicate and function symbols $Z$), called the *default formula*:

$$Z \preceq_{Di} Z' \stackrel{\text{def}}{\equiv} Di[Z] \leq Di[Z']$$

Here, $Di[Z']$ stands for the result of substituting $Z'$ for $Z$ in the formula $Di[Z]$. The $\leq$ notation is as usual in the circumscription literature. If $Di[Z]$ is a closed formula, then the right-hand-side above stands for $D[Z] \supset Di[Z']$. If $Di[Z]$ is an open formula with a tuple $x$ of individual object variables, then the right-hand-side above stands for $\forall x.\ Di[Z](x) \supset Di[Z'](x)$. *Prioritization* is defined formally as an operation that takes as input 1) a tuple of starting pre-orders; and 2) a prioritization (precedence) partial order (e.g., $R$), which is a well-founded (e.g., finite) strict partial order. The prioritization operation outputs a single, aggregated output pre-order.

A *layered* strict partial order is one that has a structure similar to the system of rank in the military. Viewed as a dag, a layered partial order consists of a totally-ordered series of one or more levels. At each level, there are one or more elements, with no links between them (i.e., within that level). Each element in a higher level has higher priority than every element in any lower level. By contrast, the typical kinds of prioritization partial orders needed to represent even simple cases of specificity in default inheritance are non-layered, e.g., *columnar*: a forest of two or more chains, with no links between the chains. Among the elements within each chain, there is a total ordering.

A prioritized *default* pre-order, e.g., $(D; R)$, is one in which each of the starting pre-orders, e.g., $Di$ (from tuple $D$), is a (single) default pre-order. A prioritized default *circumscription* is a circumscription in which the overall preference pre-order is a prioritized default pre-order. We write it as follows:

$$PDC(B; D; R; Z) \stackrel{\text{def}}{\equiv}$$
$$B[Z] \land \neg \exists Z'.\ B[Z'] \land (Z \preceq_{(D;R)} Z')$$

Here, $B$ is the base, i.e., the conjunction of all the for-sure premises (which are classical, e.g., first-order, formulas). $D$ is the tuple of (starting) default formulas, indexed say by $N$. $R$ is the prioritization partial order, defined over the tuple $N$. $Z$ is the tuple of all predicate and function symbols in the first-order logical language within which $B$ and $D$ (and $F$ below) are expressed.

When each $Di$ in $D$ is closed (i.e., propositional), then

$$Z \preceq_{(D^N;R)} Z' \stackrel{\text{def}}{\equiv}$$
$$\forall i \in N.\ [\forall j \in N.\ R(j,i) \supset (Dj[Z] \equiv Dj[Z'])]$$
$$\supset (Di[Z] \supset Di[Z'])$$

where $R(j, i)$ means that index $j$ has *higher priority* than index $i$.

In addition, we permit (explicit) *fixtures* (e.g., fixed predicates or functions):

$$PDC(B; D; R; fix\ F; Z) \stackrel{\text{def}}{\equiv}$$
$$B[Z] \land \neg \exists Z'.\ B[Z'] \land (Z \preceq_{(D;R)} Z') \land (Z \approx_F Z')$$

Here, $F$ is a tuple, indexed say by $M$, of first-order *fixture* formulas $Fk$. E.g., the fixture formula $P(x)$ expresses the fixing of the predicate $P$. E.g., the fixture formula $f(x) = y$ expresses the fixing of the function $f$.

$$Z \approx_F Z' \stackrel{\text{def}}{\equiv} \bigwedge_{k \in M} (Z \approx_{Fk} Z')$$
$$Z \approx_{Fk} Z' \stackrel{\text{def}}{\equiv} (Z \preceq_{Fk} Z') \land (Z' \preceq_{Fk} Z)$$

Fixtures can be expressed equivalently (implicitly) via default formulas: adding the fixture of the formula $Fk$ is equivalent to adding the pair of defaults $Fk$ and $\neg Fk$ in parallel to (i.e., without strict prioritization relative to) the rest of the defaults.

Equivalently, prioritized default circumscription can be defined in terms of preferences over models.

A prioritized *predicate* pre-order or circumscription can be defined as a prioritized default pre-order or circumscription, respectively, in which every default (and fixture) formula is a negated unbound atom. Minimizing a predicate, say $P$, is just a special case of a default. It corresponds to maximizing the default formula $\neg P(x)$, where $x$ is a tuple of free individual variables and has the arity of $P$.

## 3   REPRESENTING LOGIC PROGRAMS AND DEFAULT INHERITANCE IN PRIORITIZED DEFAULT CIRCUMSCRIPTION

Stratified logic programs with negation-as-failure are previously known to be representable equivalently as layered-priority predicate circumscriptions [Lifschitz, 1987] [Lifschitz, 1988] [Przymusinski, 1988]. For any stratified logic program, there is an equivalent circumscription. In the circumscription, every predicate is minimized. Every clause of the logic program is treated as a for-sure premise. The prioritization partial order corresponds to recursive depth in negation-as-failure use.

Default inheritance theories are previously known often to be representable as prioritized default circumscriptions. A straightforward method is to represent each default in the inheritance theory as one default in the circumscription, and to represent the specificity dominance partial order among defaults as the prioritization partial order. Higher specificity corresponds to higher prioritization. McCarthy's [1986] original example for prioritized circumscription is birds flying, for example. See [Grosof, 1992b] (especially, chapters 1, 2, 5, and 6) for discussion and many more examples.

More generally, one might also have other bases for prioritization-type precedence information, besides specificity, among a set of defaults. Being able to represent such is an expressive advantage of prioritized circumscription over default inheritance formalisms.

Prioritized default circumscription enables the expressive extension of default inheritance (cf. [Touretzky, 1986]) in several directions. One direction is to represent prioritization information based other than on specificity. Another direction is to represent negation, non-unary predicates, or arbitrarily nested connectives and quantifiers, in the consequent or antecedent subformulas of default "rules".

## 4   TRANSFORM EQUIVALENCE RESULTS, GENERAL CASE

Our main results are somewhat complex to state formally. So we start by giving a few simple examples to give the flavor and some intuition.

In general, the transform treats the input default formulas $E$ opaquely, and does not involve any inferencing in the sense of computing classical-logic entailments. Essentially, the transform depends only on the details of the prioritization partial order $R$, not on the form or details of the input default formulas $E$.

### Example 1 (Two Defaults)

Let two defaults' formulas be denoted by 1,2. Let the prioritization be: 1 higher than 2. Then the result of the transform is 3 parallel defaults, with formulas:
1 ,
2 $\wedge$ 1 , 2 $\vee$ 1

### Example 2 (Two Columns of Two Each)

Let four defaults' formulas be denoted by 1,2,3,4. Let the prioritization be: 1 higher than 2, and 3 higher than 4. Then the result of the transform is 6 parallel defaults, with formulas:
1 ,
2 $\wedge$ 1 , 2 $\vee$ 1 ,
3 ,
4 $\wedge$ 3 , 4 $\vee$ 3

### Example 3 (One Higher Than Two Others)

Let three defaults' formulas be denoted by 1,2,3. Let the prioritization be: 1 higher than 2, and 1 higher than 3. Then the result of the transform is 5 parallel defaults, with formulas:
1 ,
2 $\wedge$ 1 , 2 $\vee$ 1 ,
3 $\wedge$ 1 , 3 $\vee$ 1

### Example 4 (Chain of Three)

Let three defaults' formulas be denoted by 1,2,3. Let the prioritization be total: 1 higher than 2, and 2 higher than 3. Then the result of the transform is 7 parallel defaults, with formulas:
1 ,
2 $\wedge$ 1 , 2 $\vee$ 1 ,
3 $\wedge$ 2 $\wedge$ 1 , (3 $\wedge$ 2) $\vee$ 1 , (3 $\vee$ 2) $\wedge$ 1 , 3 $\vee$ 2 $\vee$ 1

### Example 5 (Two Higher than One)

Let three defaults' formulas be denoted by 1,2,3. Let the prioritization be: 1 higher than 3, and 2 higher than 3. Then one (non-deterministic) result of the transform is 7 parallel defaults, with formulas:
1 ,
2 ,
3 $\wedge$ 2 $\wedge$ 1 , (3 $\wedge$ 2) $\vee$ 1 , (3 $\vee$ 2) $\wedge$ 1 , 3 $\vee$ 2 $\vee$ 1

There is another, equivalent (non-deterministic) alternative:
1 ,
2 ,
3 $\wedge$ 1 $\wedge$ 2 , (3 $\wedge$ 1) $\vee$ 2 , (3 $\vee$ 1) $\wedge$ 2 , 3 $\vee$ 1 $\vee$ 2

### Definition 6 (General Transform to Parallel)

Let $E^N$ be a finite tuple of propositions (i.e., closed first-order formulas). (The tuple $N$ indexes $E$.)
Let $R$ be any prioritization partial order defined over $N$.

We define $\mathcal{G}$, the **general-case transform** for eliminating prioritization, as follows. $\mathcal{G}$ is a functional that depends only on the prioritization partial order $R$. $\mathcal{G}$, moreover, is non-deterministic in general. Hence, we define it as a *multi*-functional: $\mathcal{G}_R$ maps its argument $E$ into a non-empty *set*, each of whose members is a tuple of propositions.

Let $W$ be a tuple of propositions. We define $W$ to be a member of $\mathcal{G}_R(E)$ when $W$ is constructed (non-deterministically) as follows.

For each $i \in N$:
Let $\sigma^i$ be *any* sequencing of $R_D(i)$ that is descending with respect to $R$.
( $R_D(i)$ stands for the set of indices in $N$ that are higher-priority than $i$, i.e., the Dominators of $i$. "Descending" means topologically sorted in the downward direction of priority. In more detail, it means the following. Consider comparing the $i^{th}$ and the $j^{th}$ elements in the sequence, where $i < j$. Then either the (earlier) $i^{th}$ element has higher priority than the (later) $j^{th}$ element, or else the two are incomparable with respect to the prioritization partial order (neither has higher priority than the other). )

Note that since the result of topological sorting is not always unique, the choice of $\sigma^i$ is *non-deterministic* in general.

Let $m_i$ stand for the size of $R_D(i)$, and thus the length of $\sigma^i$.

Let $\sigma_k^i$ stand for the $k^{th}$ element in the sequence $\sigma^i$.

Consider the set $\mathcal{V}_i$ of all bit strings of length $m_i$. There are $2^{m_i}$ of these, corresponding to the binary numbers $\{0, \ldots, 2^{m_i} - 1\}$.

For each $l \in \mathcal{V}_i$:
Let $l_k$ stand for the $k^{th}$ bit of $l$.
Let $\gamma_k^l$ be defined as the logical connective "$\wedge$" (i.e., logical and) when $l_k = 1$, and as the logical connective "$\vee$" (i.e., logical or) when $l_k = 0$.
Let $W_{il}$ be defined as the proposition
$(E_{\sigma_1^i} \gamma_1^l (E_{\sigma_2^i} \gamma_2^l (E_{\sigma_3^i} \gamma_3^l (\ldots (E_{\sigma_{m_i}^i} \gamma_{m_i}^l E_i)\ldots))))$
where, for any $j \in N$, $E_j$ stands for the $j^{th}$ default formula in the tuple $E^N$.

Finally:
Let $W_i$ be the tuple of all the $W_{il}$'s, for all $l \in \mathcal{V}_i$.
Let $W$ be the tuple formed by concatenating (union'ing) all the $W_i$'s, for all $i \in N$.

**Theorem 7 (General Transform, Pre-Orders)**

Suppose $E^N$ is a finite tuple of *propositions*. (The tuple $N$ indexes $E$.)
Let $R$ be any prioritization partial order defined over $N$. Then the *prioritized* propositional default pre-order $(E^N; R)$ is equivalent to the *parallel* propositional default pre-order $(W; \emptyset)$, for every $W$ such that $W \in \mathcal{G}_R(E)$. ($\emptyset$ stands for the empty prioritization partial order.)

**Proof Overview:** Involved; inductive on the prioritization partial order $R$. See the longer version of this paper. □

**Proof of Example 1**:

The proof for Example 1 gives some of the flavor of the full proof, which is *much* more complicated.

Let $i'$ for $i = 1, 2$ stand for the default formula $i$ with $Z'$ substituted for $Z$. We want to show that
$(1 \supset 1') \wedge ((1 \wedge 2) \supset (1' \wedge 2'))$
$\wedge ((1 \vee 2) \supset (1' \vee 2'))$
is equivalent to
$(1 \supset 1') \wedge ((1 \equiv 1') \supset (2 \supset 2'))$
Assume:

$(1 \supset 1')$ \hfill (A1)

Then it suffices to show

$((1 \wedge 2) \supset (1' \wedge 2')) \wedge ((1 \vee 2) \supset (1' \vee 2'))$ \hfill (2)

is equivalent to

$((1 \equiv 1') \supset (2 \supset 2'))$ \hfill (3)

(A1) implies that (2) is equivalent to

$((1 \wedge 2) \supset 2') \wedge (2 \supset (1' \vee 2'))$ \hfill (4)

(4) is tautologically equivalent to

$(1 \supset (2 \supset 2')) \wedge (\neg 1' \supset (2 \supset 2'))$ \hfill (5)

(A1) implies tautologically that

$(1 \equiv (1 \wedge 1')) \wedge (\neg 1' \equiv (\neg 1 \wedge \neg 1'))$ \hfill (6)

(6) implies by substitutional rewriting that (5) is equivalent to

$((1 \wedge 1') \supset (2 \supset 2')) \wedge ((\neg 1 \wedge \neg 1') \supset (2 \supset 2'))$ \hfill (7)

(7) is tautologically equivalent to

$((1 \wedge 1') \vee (\neg 1 \wedge \neg 1')) \supset (2 \supset 2')$ \hfill (8)

Tautologically, the left-hand-side of (8) is equivalent to the left-hand-side of (3).

**QED Example 1**

**Theorem 8 (Transform, Circumscriptions)**

Let $PDC(B; E; R; Z)$ be any *prioritized* default circumscription defined without (explicit) fixtures. In particular, $R$ may be any arbitrary prioritization partial order. (E.g., $R$ need not be layered.)

Suppose the default formulas $E$ are all propositional. Then the PDC is equivalent to the parallel default circumscription that results from applying the transform $\mathcal{G}$, i.e.:
$$PDC(B; E; R; Z) \equiv PDC(B; W; \emptyset; Z)$$
for every $W \in \mathcal{G}_R(E)$.

Matters are similar for the case when any (explicit) fixtures $F$ are present:
$$PDC(B; E; R; fix\ F; Z) \equiv PDC(B; W; \emptyset; fix\ F; Z)$$

**Proof**: We begin by considering the case without (explicit) fixtures:

We are comparing two circumscriptions, one before the transform and one after the transform. They have the same base $B$. The definition of circumscription implies, therefore, that two circumscriptions are equivalent if their preference pre-orders (($E; R$) and ($W; \emptyset$), respectively) are equivalent. Theorem 7 implies just such equivalence between the preference pre-orders.

That adding (explicit) fixtures preserves the equivalence is seen easily by inspecting the role of fixture in the definition of circumscription. □

**Remarks:**

The base $B$ and the fixtures $F$ above need not be propositional.

The equivalence between the prioritized set of defaults $(E; R)$ and the parallel set of defaults $(W; \emptyset)$ is strong,

in the sense that it holds for *any* base (or fixture). Thus if there are a series of updates to the base (or fixtures) alone, then the transform does not need to be re-applied.

The non-determinism in the definition of the transform does not increase required computational effort. All it does is provide equivalent alternatives.

**Observation 9 (Effective Propositionality)**

When the base implies
1. domain closure (DCA); plus
2. uniqueness of names (UNA) (or, more generally, a complete theory of equality),

then the defaults in a circumscription, and in its pre-orders, are effectively propositional, in the sense that every default can be viewed as equivalent to the (parallel) collection of all its ground instances. More precisely: in a PDC, when the base implies domain closure plus uniqueness of names, then this first PDC is equivalent to a second PDC defined by replacing every default in the first PDC by the collection of its instances.

**Proof**: See [Grosof, 1992b]: Theorem 2.47. □

**Theorem 10 (Extension to DCA & UNA)**

Our results in Theorem 8 thus generalize straightforwardly to permit the default formulas to be *open or closed*, when the restriction in Observation 9 is met. Convert the initial set of prioritized defaults to a set of prioritized closed defaults, by replacing every default by the collection of its instances. Then apply Theorem 8.

**Proof**: Immediate from Observation 9. □

## 5 ALGORITHM TO IMPLEMENT THE GENERAL TRANSFORM

Implementing the general-case transform $\mathcal{G}$ is straightforward. Next, we sketch an algorithm to compute $\mathcal{G}_R(E)$:

1. For each $i \in N$:
2. compute $R_D(i)$ and its size $m_i$;
3. non-deterministically, topologically sort $R_D(i)$, resulting in $\sigma^i$;

As one simple way to compute step 3., one might do the following:
Before step 1. above: non-deterministically, topologically sort $R$, resulting in a sequence that we may call $\psi$. Then, in step 3., compute $\sigma^i$ as the intersection of $\psi$ with $R_D(i)$.

4. construct $\mathcal{V}_i$;
5. For each $l \in \mathcal{V}_i$:

Alternatively, one might skip step 4. and instead generate $\mathcal{V}_i$ within 5. as one goes.

6. construct $W_{il}$: first, initialize it to be $E_i$, then:

7. For each $k = m_i, \ldots, 1$:
8. add another $E_{\sigma_k^i} \gamma_k^l$ then parenthesize.

9. Finally, collect all the $W_{il}$'s: the result $W$ is, as desired, a member of $\mathcal{G}_R(E)$.

**Computational Complexity of the General Transform:**
Worst case, the transform is exponential in the size of the input representation of $E$ and $R$, i.e., in the number $n$ of input (prioritized) defaults. The nub is the size of $W$ itself, which is worst-case exponential in $n$.

However, when the size of $W$ is polynomial, then the time required to compute the transform is also polynomial. This is the case, for example, when: for each default $i$ in $E^N$, the size $m_i$ of $R_D(i)$ is bounded by a constant, or grows slowly, say is $O(\log n)$. Recall that $R_D(i)$ is the set of defaults that are higher-priority than default $i$. Thus if the prioritization partial order $R$ is non-"**top-heavy**" in this sense, then the transform is tractable.

**Layered or Total Restrictions Do Not Help**: We observe that the (size) complexity of the transform is not, in general, improved by requiring that the prioritization partial order be layered, nor by requiring it to be totally ordered. A total order is always top-heavy. Layered partial orders are often top-heavy.

## 6 SPECIAL CASE TRANSFORM FOR SPECIFICITY IN INHERITANCE

The general-case transform $\mathcal{G}$ can also be used analytically to develop special-case transforms that are much simpler than the general-case transform.

In particular, it can be used to develop a special-case transform for default inheritance, where prioritization is based on specificity, which is in turn based on for-sure beliefs about a taxonomic hierarchy. Preliminary investigations indicate that this specificity transform has only quadratic (often linear) blow-up in the number/size of defaults, and thus polynomial computational time complexity. Due to space limitations here, we discuss the details and general results elsewhere in a forthcoming paper. To give the flavor, however, next we give some examples.

All of these examples are about inheritance by a single individual, let us call her *Tweety*, of a single attribute, flying; however, they straightforwardly extend to a full domain of many individuals, and inheritance of many attributes. Below, for conciseness of exposition, we leave *Tweety* implicit and omit showing it as an argument of the predicates *bird*, *flies*, etc..

**Example 11 (One Exceptional Sub-Class)**

Let the starting representation with priorities be:

$E1 \stackrel{\text{def}}{=} bird \supset flies$

$$E2 \stackrel{def}{\equiv} ostrich \supset \neg flies$$

where $E2$ has higher priority than $E1$, and the base/for-sure axioms entail that $ostrich \supset bird$. This is a very simple case of an default inheritance chain. Then after the special-case transform, there are two defaults, with formulas:

$$D1 \stackrel{def}{\equiv} bird \supset (flies \wedge \neg ostrich)$$
$$D2 \stackrel{def}{\equiv} ostrich \supset \neg flies$$

**Example 12 (Two Exceptional Sub-Classes)**

Suppose we extend Example 11 by adding another default:

$$E3 \stackrel{def}{\equiv} penguin \supset \neg flies$$

where $E3$ has higher priority than $E1$, and the base/for-sure axioms entail that $penguin \supset bird$. Then after the special-case transform, there are three defaults, with formulas:

$$D1 \stackrel{def}{\equiv} bird \supset (flies \wedge \neg ostrich \wedge \neg penguin)$$
$$D2 \stackrel{def}{\equiv} ostrich \supset \neg flies$$
$$D3 \stackrel{def}{\equiv} penguin \supset \neg flies$$

**Example 13**

**(Two Levels of Exceptional Sub-Classes)**

Suppose we extend Example 12 by adding another default:

$$E0 \stackrel{def}{\equiv} animal \supset \neg flies$$

where $E1$ (and $E2$ and $E3$) has higher priority than $E0$, and the base/for-sure axioms entail that $bird \supset animal$. Then after the special-case transform, there are four defaults, with formulas:

$$D0 \stackrel{def}{\equiv} animal \supset (\neg flies \wedge \neg bird)$$
$$D1 \stackrel{def}{\equiv} bird \supset (flies \wedge \neg ostrich \wedge \neg penguin)$$
$$D2 \stackrel{def}{\equiv} ostrich \supset \neg flies$$
$$D3 \stackrel{def}{\equiv} penguin \supset \neg flies$$

**Examples Proof Overview:** Our proof technique for the above Examples, and for the special case transform more generally, relies on results in [Grosof, 1992b]. The special-case transform, e.g., in the examples above, has as output a *subset* of the defaults that result from the general-case transform. Defaults can be dropped (from the general-case's output) when they are *redundant* in the sense of [Grosof, 1992b]'s Definition 4.17 and Theorem 6.29. A default is redundant if the base implies 1) that its formula $Wil$ is equivalent to an expression formed positively (i.e., using $\wedge$, $\vee$, and quantifiers but not $\neg$) from other default formulas (in $W$); or 2) that $Wil$ is tautologically true; or 3) that $Wil$ is tautologically false. □

**Relationship to Parallel Abnormalities With Cancellation**: There is a close relationship between the above examples and a well-known method of employing abnormality predicates to represent a simple default inheritance chain: by minimizing the abnormality predicates, with explicit cancellation axioms, and with *parallel prioritization*. For the above examples, the following defines an equivalent representation. Introduce one distinct abnormality predicate $abi$ for each default formula $Di$ above. Assert a for-sure premise for each $i$:

$$\neg abi \supset Di$$

Assert *explicit cancellation axioms* as for-sure premises, as follows. For each pair $(j,i)$ of defaults such that $j$ has higher priority (is more specific) than $i$, include either the for-sure premise:

$$\neg Dj \supset abi$$

or, alternatively (equivalently), the for-sure premise:

$$\neg cj \supset abi$$

where $cj$ is the class condition on the left-hand-side of the rule $Dj$, e.g., $ostrich$ in the left-hand-side of $D2$. Finally, minimize all the abnormality predicates in parallel. (No other defaults are maximized.)

**Proof Overview for Equivalence of Examples to Abnormalities plus Cancellation**: The equivalence of this representation, for the above examples, can be proven straightforwardly using the results on abnormality theories, and their equivalence relationship to maximizing default formulas, in [Grosof, 1992b] (section 3.2 there). □

Thus abnormalities with explicit cancellation provides an alternative way to transform a prioritized default representation into a parallel one, for these examples.

## 7 USES FOR IMPLEMENTATION

The transforms, both general-case and special-case, are useful as a front-end to inferencing algorithms that handle the parallel expressive class.

As we discussed earlier (recall Theorem 10), our general-case transform requires only one expressive restriction: that 1) the defaults are propositional; or 2) the base implies domain closure and uniqueness of names / complete theory of equality. In case of 2), recall, the defaults are effectively propositional.

### 7.1 REVIEW OF PREVIOUSLY AVAILABLE INFERENCING ALGORITHMS

There are several previously available query-answering inferencing algorithms for parallel defaults in circumscription that are sound and complete for queries over fairly general expressive classes of first-order-form beliefs: [Przymusinski, 1989] [Ginsberg, 1989] [Baker and Ginsberg, 1989] [Inoue and Helft, 1990] [Helft et al., 1991]. Each of these covers (at least) ground queries. All of these essentially require effective propositionality. Each, besides Przymusinski's, covers (at least) the expressive class of: domain closure plus uniqueness of names, which includes propositionality (of base and defaults) as a special case, of course. All restrict the default formulas to have the form of minimizing a predicate: i.e., every default formula $Di$ has the form of a negated atom $\neg Pi(x)$, where $Pi$ is a predicate symbol and $x$ is a tuple of free variables. However, this restriction is inessential. Maximizing arbitrary default formulas can be reduced, in a simple fashion, to minimizing predicates. [Grosof, 1992b] (section 7.6 there) shows a general method, imposing only overhead time that is polynomial in the input representation, to convert inferencing with arbitrary default formulas (and fixtures) to inferencing with default formulas that are restricted to be predicate minimizations (with all fixtures expressed as fixing of predicates or functions). This method requires no other restrictions on the (input) base, priorities, fixtures, or form of conclusions. This method is based on introducing abnormality predicates. It extends and refines the by-now well-known style of abnormality theories, so as to apply to the case of arbitrary prioritization partial orders. In addition, the algorithms of [Ginsberg, 1989] and [Baker and Ginsberg, 1989] are essentially developed, conceptually as well as mathematically, in terms of maximizing default pre-orders. Thus they are easily extended to handle correctly the default, not just the predicate, case.

The algorithms of [Ginsberg, 1989] and [Baker and Ginsberg, 1989] enable arbitrary closed queries, but impose the restriction of no fixed predicates; however, [Inoue and Helft, 1990] shows how to relax this restriction to permit arbitrary fixed predicates. [Helft et al., 1991] shows how to extend to queries with answer extraction, not just closed (yes/no) queries.

The punchline is that previously available algorithms support rich query-answering for parallel defaults in circumscription, for the expressive class of domain closure plus uniqueness of names (which includes propositional).

In addition, two of the previously available query-answering algorithms, [Baker and Ginsberg, 1989] and [Przymusinski, 1989] (who does not give a proof), apply to the prioritized case, *but* require that the prioritization partial order be **layered**. (Both also require queries to be closed.)

### 7.2    NEW INFERENCE ALGORITHMS

Our general-case transform enables the extension of any parallel-case algorithm $\mathcal{IP}$ to handle defaults with **arbitrary** prioritization. These include all of the previous algorithms discussed above, as well as any other / future algorithms. Our general-case transform applies to any direction of inferencing: e.g., the direction may be backward (query-answering) or forward (exhaustive; or selective, e.g., data-driven); queries may be ground, closed, or open (answer extraction). And it applies to inferencing algorithms for expressively highly restricted parallel cases, as well as to those for expressively general parallel cases. Special case transforms based on $\mathcal{G}$, e.g., the one we briefly discussed in section 6, apply similarly, though of course with appropriate expressive restrictions.

The method is simple:
1. If the input prioritized defaults are not propositional, but there is domain closure and uniqueness of names / complete theory of equality, then replace every (open) default by the collection of its instances. In practice, this need not be computed explicitly in its entirety, e.g., for defaults about which there is no strict prioritization information.
2. Apply the transform to the input prioritized defaults. This results in a representation where the defaults are parallel.
3. Apply the inference procedure $\mathcal{IP}$.

For example, the [Inoue and Helft, 1990] algorithm can thus be extended to cover the expressive case of: domain closure plus uniqueness of names, for arbitrary prioritization, universal default formulas, universal base, quantifier-free fixtures, and closed queries. This is a fairly expressive class, and compares well with the expressive classes used practically today in monotonic first-order-logic inferencing.

Another potential use of our transform approach is to *partially* eliminate the input's prioritization, so as to produce a simpler prioritization, e.g., layered, in the output that can be handled by another available algorithm, e.g., [Baker and Ginsberg, 1989]'s.

## 8    COMPUTATIONAL PRACTICALITY

### 8.1    COMPUTATIONAL COMPLEXITY OF INFERENCING, GENERALLY, INCLUDING WITHOUT PRIORITIES

A major practical difficulty for default reasoning, even without priorities, is the worst-case computational complexity of inferencing. Skeptical entailment (i.e., answering a single closed query) is known to be $\Pi_2^P$-complete, i.e., co-NP-harder than monotonic entailment, in the propositional case of several expressively rich non-monotonic logical formalisms, including: circumscription (even minimizing predicates without priorities), Default Logic (even the "normal" case), Autoepistemic Logic, and several additional non-monotonic modal logics and other formalisms [Eiter and Gottlob, 1993] [Stillman, 1992] [Gottlob, 1992].

There are several avenues to avoiding worst-case complexity of default reasoning, generally. One is to employ approximations, perhaps sound but incomplete, e.g., as in [Cadoli and Schaerf, 1992]. Another is to restrict expressive classes to those with significantly better complexity, e.g., such that inferencing over a large theory can be decomposed into inferencing over several smaller or simpler theories. [Grosof, 1992b] explores both of these avenues, especially the latter.

### 8.2    COMPUTATIONAL PRACTICALITY OF INFERENCING IMPLEMENTED VIA OUR METHOD

Our transform, as used in our inference algorithms for prioritized defaults, potentially compounds the computational complexity situation for parallel defaults with yet another source of worst-case exponential complexity: in time, and in the number of defaults input into parallel-case inferencing. Thus, it must be used with some discretion, just as parallel default reasoning must be in general.

Implementing the transform is practical nevertheless in two kinds of situations: general expressiveness but small numbers of defaults, or expressive special cases with larger numbers of defaults.

First, our method enables the implementation at least of small numbers of prioritized defaults that have arbitrary non-layered prioritization partial orders. The history of rule-based knowledge representation demonstrates that even small rule sets are often quite useful for applications.

Second, feasibility for small numbers of defaults can be sometimes be **leveraged** in order to implement larger prioritized default theories. There exist previous techniques for **decomposing** large prioritized default theories **into a collection of local** default theories, each with a small number of defaults [Grosof, 1992b] [Grosof, 1992a].

Third, complexity of the transform can be kept manageable by **expressively limiting** the **top-heaviness** ($m_i$) of the prioritization partial order. An example is when the (strict) prioritization is columnar (recall section 2 terminology) and the columns (chains) have bounded height (e.g., less than five). Shallowness of rule interaction is a common situation in practical knowledge-based systems today.

Fourth, complexity of the transform can be kept manageable by employing special-case transforms, corresponding to restricted expressive classes, e.g., the tractable special-case transform for specificity in inheritance that we briefly discussed in section 6.

Finally, we observe also that the transform is a "compile-time" operation that need be performed only once for a long sequence of for-sure (base) updates and even default updates, as long as the (strict) prioritization is not changed.

The computational complexity of the overall inferencing does not depend only on the computational complexity of the transform. It also depends on the underlying computational complexity of inferencing with the parallel defaults that result from the transform. As we discussed in subsection 8.1, this itself is $\Pi_2^P$-complete.

Again, the worst-case complexity of inferencing with parallel defaults limits, but far from eliminates, the practicality of such inferencing. In some interesting and useful cases, parallel default reasoning is tractable. Again, there are two kind of situations. The first kind is general expressiveness but small numbers of defaults. The second kind is expressive special cases with larger numbers of defaults: e.g., Closed World Assumption, some kinds of logic programs and predicate completions, sympathetically solitary default theories [Grosof, 1992b] (section 6.5 there), and some other cases.

Let us summarize the foregoing's implications for overall inferencing via our method, combining both the transform and the parallel reasoning. The feasibility picture is of a glass half full and half empty, just as in much of AI and KR. Inferencing via our method is computationally practical for small numbers of defaults with general expressiveness, or for larger numbers of defaults in expressive special cases for which the underlying default reasoning *and* the transform are tractable.

## 9  MORE IMPLICATIONS OF FUNDAMENTAL RESULT

### 9.1  PRIORITIZATION'S EXPRESSIVE REDUCIBILITY AND CONCISENESS

One can view the transform as reducing $n$ degrees of partially ordered belief confidence to just 2 degrees of confidence: for-sure and (unprioritized) default.

This reducibility is, at first glance, surprising: prioritization was introduced into circumscription by McCarthy [1986], Lifschitz [1985], and Grosof [1991] because it was not understood how to achieve the same entailment behavior without it; likewise, for similar concepts of precedence in other non-monotonic formalisms.

At second glance, however, the reducibility is less surprising. Ordinary, parallel default reasoning, e.g., in parallel circumscription or Poole's Theorist, can be viewed in these terms as reducing 2 degrees of confidence to just 1 degree of confidence: that of the non-monotonic theory's conclusions.

Much of the point of non-monotonic reasoning (e.g., $\Pi_2^P$-complete; recall subsection 8.1) altogether is its representational conciseness (e.g., exponential) relative to monotonic reasoning (e.g., NP-complete).

Nevertheless, prioritization appears to be useful as a tool for conciseness, and thus naturalness, in representation. Our transform result suggests (no lower-bound result, though) that prioritization may allow an exponential savings in the number of defaults that must be specified by a user.

In short, the expressive reduction's computational complexity suggests that prioritization is valuable for its expressive conciseness, just as defaults are for theirs.

### 9.2  DEFINING PRIORITIZATION IN FORMALISMS PREVIOUSLY WITHOUT IT

Our transform implies how to define / introduce explicit prioritization, in a manner precisely equivalent to prioritization in default circumscription, into some previously parallel formalisms. In particular, Theorist Poole's formalism [1988] and the "normal, prerequisite-free" case of Reiter's [1980] Default Logic each overlap equivalently with parallel fixture-free default circumscription, in the case of propositionality or domain closure plus uniqueness of names.

In a bit more detail: Grosof [1992b] (section 8.5 there) shows that, in this case, for any set of default formulas $G$, indexed by $N$:
the parallel default circumscription without explicit fix-tures $PDC(B; G^N; \emptyset; Z)$
is equivalent to the skeptical version of the Default Logic default theory $\langle B, D \rangle$,
where each DL default $Di \in D$, for $i \in N$, is defined as:

$$\frac{: Gi}{Gi}$$

Thus, our transform can be applied to define prioritized versions of Poole's Theorist formalism and of normal, prerequisite-free Default Logic. It will be interesting to compare this to Brewka [1989a] [1989b] [1994] alternative approach in which he defines new extended, prioritized variants of Poole's Theorist formalism and of Default Logic. Brewka's approach does not, in general, have the same entailment behavior (see [Grosof, 1992b] section 8.8).

## 10 CONTRAST WITH OTHER ALGORITHMS AND APPROACHES

Our transformational approach (for transforming prioritized into parallel) and algorithms (for inferencing with prioritization) is quite different from previous approaches and algorithms for circumscription. Perhaps most importantly, unlike previous ones, it extends to non-layered prioritization partial orders, which appear quite important for practical applications; e.g., default inheritance typically has non-layered prioritization.

Baker & Ginsberg's [1989]'s query-answering algorithm for the layered case of prioritized default circumscription does not reduce the prioritized representation to a parallel one; instead it takes prioritization into account by modifying the dominance criterion involved in dialectically comparing arguments for and against a given proposition.

The previous approach to transforming a prioritized set of defaults into a parallel set of defaults is based on what [Grosof, 1992b] calls *decomposition*. [Lifschitz, 1985] has shown an equivalence theorem, for the case of layered prioritization and predicate minimization, between a prioritized circumscription and a conjunction of parallel circumscriptions, one per layer. [Grosof, 1992b] (chapters 5 and 7, especially) has extended this to the case of arbitrary non-layered prioritization and arbitrary default formulas, and has shown a similar result giving equivalence to a *series* (cascade) of parallel circumscriptions (extending a previous unpublished result by Lifschitz for the layered predicate case). In addition, Brewka's [1989a] [1989b] [1994] employs the serial decompositional approach as a means to definitionally introduce prioritization-type precedence into Poole's [1988] Theorist formalism and into Default Logic.

The decompositional approach is quite different in spirit from our transform here, which gives equivalence to a single parallel circumscription. The decompositional approach, in general, results in extra fixtures and also, for non-layered prioritizations, expressively complex extra base. Also, neither Grosof nor Lifschitz have previously given a query-answering algorithm based on this decompositional approach. (Though [Przymusinski, 1989]'s unproven algorithm for layered predicate case is based on Lifschitz' result.)

Besides previous work for general prioritized circumscriptions, there is also some relevant previous work on encoding specificity into parallel defaults. Early work by Etherington & Reiter [1983] showed how to encode specificity in inheritance into Default Logic, starting from a path-based representation cf. [Touretzky, 1986]. Delgrande & Schaub [1994] propose a general approach, applicable to many default formalisms: "to use the techniques of a weak system, as exemplified by System Z [ [Pearl, 1990] ], to isolate minimal sets of conflicting defaults. From the specificity information intrinsic in these sets, a default theory in a target language is specified." They then give some particular transforms, concentrating primarily on Default Logic. In current work, they are exploring applying their approach to circumscription.

It will be interesting to compare the transformational approach here in more detail to the previous approaches, in the cases where they overlap: e.g., to compare their entailment behavior and the efficiency of inferencing and updating algorithms based on the different approaches.

## 11 SUMMARY; CURRENT AND FUTURE WORK

**SUMMARY OF PAPER**: See the Abstract.

**MORE EFFICIENT SPECIAL CASE FOR SPECIFICITY AND INHERITANCE**: Recall section 6 which described a forthcoming paper.

**APPLYING RESULTS TO OTHER FORMALISMS**: In a forthcoming paper, we also show that our results apply to several default formalisms other than circumscription, including [Geffner, 1992]'s formalism for defaults, inheritance, specificity, and conditionals; [Quantz and Royer, 1992]'s formalism for defaults, inheritance, and specificity in terminological logics; and Brewka's formalism family [1989a] [1989b] [1994] which extends Poole's Theorist and Reiter's Default Logic with prioritization-type precedence.

Each of these formalisms overlap equivalently, for a broad case, with propositional prioritized defaults in circumscription. Our results thus apply to these formalisms as well.

**REMAINING CHALLENGES**: We have not yet experimented with, or evaluated the efficiency of, our approach and algorithms: that awaits future work.

There is much further to go to realize the vision we discussed towards the end of section 1. Enabling sufficient computational *efficiency* and demonstrating practical applications remain outstanding as problems for future work.

Good algorithms for other kinds of reasoning besides query-answering are also needed: for updating and belief revision, and for other directions, e.g., forward, of inferencing. Approaches to these are discussed for prioritized default circumscription in [Grosof, 1992b] [Grosof, 1992a].

## Acknowledgements

Thanks to Torsten Schaub, Jim Delgrande, Leora Morgenstern, and Hector Geffner for encouraging discussions. Thanks also to two anonymous reviewers for helpful suggestions, some of which are incorporated only in the longer version of this paper.